\documentclass[twoside,11pt]{article}

\usepackage{jmlr2e}
\usepackage{natbib}

\usepackage{algorithm}
\usepackage{algorithmic}
\usepackage{amssymb,amsmath}

\renewcommand{\algorithmicrequire}{\textbf{Arguments:}}
\long\def\symbolfootnote[#1]#2{\begingroup%
\def\thefootnote{\fnsymbol{footnote}}\footnote[#1]{#2}\endgroup} 

\jmlrheading{}{}{}{}{}{Vicen\c{c} G\'omez, Hilbert J. Kappen and Michael Chertkov}

\ShortHeadings{Approximate inference on planar graphs using Loop Calculus and BP}
{G\'omez, Kappen and Chertkov }
\firstpageno{1}

\begin{document}

\title{Approximate inference on planar graphs using Loop Calculus and Belief Propagation}

\author{\name Vicen\c{c} G\'omez \email v.gomez@science.ru.nl \\
\name Hilbert J. Kappen
\email b.kappen@science.ru.nl \\
\addr Department of Biophysics\\ Radboud University Nijmegen\\ 6525 EZ Nijmegen, The Netherlands\\
\AND
\name Michael Chertkov
\email chertkov@lanl.gov \\
\addr Theoretical Division and Center for Nonlinear Studies\\ Los Alamos National Laboratory\\ Los Alamos, NM 87545
}

\editor{}
\maketitle

\begin{abstract}
We introduce novel results for approximate inference on planar graphical models
using the loop calculus framework.  The loop calculus \citep{chertkov2006a}
allows to express the exact partition function of a graphical model as a finite
sum of terms that can be evaluated once the belief propagation (BP) solution is
known.  In general, full summation over all correction terms is intractable.
We develop an algorithm for the approach presented in \citet{chertkov08} which
represents an efficient truncation scheme on planar graphs and a new
representation of the series in terms of Pfaffians of matrices.  We analyze the
performance of the algorithm for the partition function approximation for
models with binary variables and pairwise interactions on grids and other planar
graphs.  We study in detail both the loop series and the equivalent
Pfaffian series and show that the first term of the Pfaffian series for the
general, intractable planar model, can provide very accurate approximations.
The algorithm outperforms previous truncation schemes of the loop series and is
competitive with other state-of-the-art methods for approximate inference.

\end{abstract}


\section{Introduction}
\label{sec:intro}
Graphical models are popular tools widely used in many areas which require
modeling of uncertainty.  They provide an effective approach through a compact
representation of the joint probability distribution.  The two most common
types of graphical models are Bayesian Networks (BN) and Markov Random Fields
(MRFs).

The partition function of a graphical model, which plays the role of
normalization constant in a MRF or probability of evidence (likelihood) in a BN
is a fundamental quantity which arises in many contexts such as hypothesis
testing or parameter estimation.  Exact computation of this quantity is only
feasible when the graph is not too complex, or equivalently, when its
tree-width is small.  Currently many methods are devoted to approximate this
quantity.

The belief propagation (BP) algorithm \citep{pearl88} is at the core of many of
these approximate inference methods.  Initially thought as an exact algorithm
for tree graphs, it is widely used as an approximation method for loopy graphs
\citep{murphy99, Frey}.  The exact partition function is explicitly related to
the BP approximation through the loop calculus framework introduced by
\citet{chertkov2006a}.  Loop calculus allows to express the exact partition
function as a finite sum of terms (loop series) that can be evaluated once the
BP solution is known.  Each term maps uniquely to a subgraph, also denoted as a
generalized loop, where the connectivity of any node within the subgraph is
\emph{at least} degree two. Summation of the entire loop series is a hard
combinatorial task since the number of generalized loops is typically
exponential in the size of the graph.  However, different approximations can be
obtained by considering different subsets of generalized loops in the graph.

It has been shown empirically \citep{gomez07, chertkov2006b} that truncating
this series may provide efficient corrections to the initial BP approximation.
More precisely, whenever BP performs satisfactorily which occurs in the case of
sufficiently weak interactions between variables or short-range influence of
loops, accounting for only a small number of terms is sufficient to recover the
exact result~\citep{gomez07}.  On the other hand, for those cases where BP
requires many iterations to converge, many terms of the series are required to
improve substantially the approximation. A formal characterization of the
classes of tractable problems via loop calculus still remains as an open
question.

A step toward this goal has been done in \citet{chertkov08} where it was shown
that for any graphical model, summation of a certain subset of terms can be
mapped to a summation of weighted perfect matchings on an extended graph.  For
planar graphs (graphs that can be embedded into a plane without crossing edges),
summation of the subset can be performed in polynomial time evaluating the Pfaffian of a
skew-symmetric matrix associated with the extended graph.  Furthermore, the full
loop series can be expressed as a sum over certain Pfaffian terms, where each
Pfaffian term accounts for a large number of loops and is solvable in
polynomial time as well.

The approach of \citet{chertkov08} builds on classical results from $1960$s by
\citet{kasteleyn63,Fisher66} and other physicists who addressed the question of
counting the number of perfect matchings on a planar grid, also known as the
dimer problem in the statistical physics literature (a dimer correspond to a
colored edge of the graph, and a valid dimer configuration consists of exactly
one dimer per any edge of the graph).  The key result of
\citet{kasteleyn63,Fisher66} can be summarized as follows: the partition
function of a \emph{planar graphical model defined in terms of binary
variables} can be mapped to a weighted perfect matching problem and calculated
in polynomial time under the restriction that interactions only depend on
agreement or disagreement between the signs of their variables.  Such a model
is known in statistical physics as the Ising model \emph{without external
field}.  Notice that exact inference for a general binary graphical model on a
planar graph (that is Ising model with external field) is intractable
\citep{Barahona82}.

Recently, some methods for inference over graphical models, based on the works
of Kasteleyn and Fisher, have been introduced. \citet{Globerson2006} obtained
upper bounds on the partition function for non-planar graphs with binary
variables by decomposition of the partition function into a weighted sum over
partition functions of spanning tractable (zero field) planar models.  The
resulting problem is a convex optimization problem and, since exact inference
can be done in each planar \emph{sub-}model, the bound can be calculated in
polynomial time.

Another example is the work of \citet{Schraudolph} which provides a framework
for exact inference on a restricted class of planar graphs using the approach
of Kasteleyn and Fisher.  More precisely, they showed that any joint
probability function defined on binary variables can be expressed in a
functional form without external fields by adding a new auxiliary node linked
to all the existing nodes.  Under this transformation, single-variable external
fields can be allowed for a subset $\mathcal{B}$ of variables.  If the
graphical model is $\mathcal{B}-$outerplanar, which means that there exists a
planar embedding in which the subset $\mathcal{B}$ of the nodes lie on the
same face, the techniques of Kasteleyn and Fisher can still be applied.

Contrary to the two aforementioned approaches which rely on exact inference on
a tractable planar model, the loop calculus directly leads to a framework for
approximate inference on general planar graphs.  Truncating the loop series
according to \citet{chertkov08} already gives the exact result in the zero
external field case.  In the general planar case, however, this truncation may
result into an accurate approximation that can be incrementally corrected by
considering subsequent terms in the series.

In the next Section we review the main theoretical results of the loop calculus
approach for planar graphs and introduce the proposed algorithm.  In
Section~\ref{sec:experim} we provide experimental results on approximation of
the partition function for regular grids and other types of planar graphs.  We
focus on a planar-intractable binary model with symmetric pairwise interactions
but nonzero single variable potentials.  The source code used to derive these
results is freely available at \verb|http://www.mbfys.ru.nl/staff/v.gomez/|.
We end this manuscript with conclusions and future work in
Section~\ref{sec:discussion}.

\section{Belief Propagation and loop Series for Planar Graphs}
\label{sec:method}
We consider the Forney graph representation, also called general vertex model
\citep{Forney, Loeliger04}, of a probability distribution
$p(\boldsymbol{\sigma})$ defined over a vector $\boldsymbol{\sigma}$ of binary
variables (vectors are denoted using bold symbols).  Forney graphs are
associated with general graphical models which subsume other factor graphs,
e.g. those correspondent to BNs and MRFs.  In Appendix A we show how to convert
a factor graph model to its equivalent Forney graph representation.

A binary Forney graph $\mathcal{G}:=(\mathcal{V},\mathcal{E})$ consists of a
set of nodes $\mathcal{V}$ where each node $a\in\mathcal{V}$ represents an
interaction and each edge $(a,b)\in\mathcal{E}$ represents a binary variable
$ab$ which take values $\sigma_{ab} :=\{\pm 1\}$.  We denote $\bar{a}$ the set
of neighbors of node $a$.  Interactions $f_a\left(\boldsymbol{\sigma}_a\right)$
are arbitrary functions defined over typically small subsets of variables where
$\boldsymbol{\sigma}_a$ is the vector of variables associated with node $a$,
i.e. $\boldsymbol{\sigma}_a := (\sigma_{ab_1}, \sigma_{ab_2}, \dots)$ where
$b_i\in\bar{a}$.

The joint probability distribution of such a model factorizes as:
\begin{align}\label{eq:model}
p\left(\boldsymbol{\sigma}\right) & =
Z^{-1}\prod_{a\in\mathcal{V}}{f_a\left(\boldsymbol{\sigma}_a\right)}, &
Z & = \sum_{\boldsymbol{\sigma}}{\prod_{a\in\mathcal{V}}{f_a\left(\boldsymbol{\sigma}_a\right)}},
\end{align}
where $Z$ is the normalization factor, also called the partition function.

From a variational perspective, a fixed point of the BP algorithm represents a
stationary point of the Bethe "free energy" approximation under proper constraints
\citep{yedidia00}.  In the Forney style notation:
\begin{align}\label{eq:Bethe}
Z^{BP} & = \exp\left(-F^{BP}\right), \\
F^{BP} & =
\sum_{a}{\sum_{\boldsymbol{\sigma}_a}{b_a\left(\boldsymbol{\sigma}_a\right)\text{ln}\left(\frac{b_a(\boldsymbol{\sigma}_a)}{f_a(\boldsymbol{\sigma}_a)}\right)}}\notag
-\sum_{b\in\bar{a}}{\sum_{\sigma_{ab}}{b_{ab}\left(\sigma_{ab}\right)\text{ln}{b_{ab}\left(\sigma_{ab}\right)}}},
\end{align}
where $b_a(\boldsymbol{\sigma}_a)$ and $b_{ab}(\sigma_{ab})$ are the beliefs
(pseudo-marginals) associated to each node $a \in \mathcal{V}$ and variable
$ab$.  For graphs without loops, Equation~\eqref{eq:Bethe} coincides with the
Gibbs "free energy" and therefore $Z^{BP}$ coincides with the exact partition
function $Z$. If the graph contains loops, $Z^{BP}$ is just an approximation critically dependent 
on how strong the influence of the loops is.

We introduce now some convenient definitions related to the loop calculus framework.
\begin{definition}\label{def:gen loop}
A \textbf{generalized loop} in a graph $\mathcal{G}=\langle \mathcal{V}, \mathcal{E}\rangle$ is any
subgraph $C = \langle V', E'\rangle$, $V' \subseteq \mathcal{V}, E' \subseteq (V'\times V') \cap \mathcal{E}$
such that each node in $V'$ has degree two or larger.
\end{definition}
For simplicity, we will use the term "loop", instead of "generalized loop", in
the rest of this manuscript.
Loop calculus allows to represent $Z$ explicitly in terms of the BP
approximation via the loop series expansion:
\begin{align}\label{eq:fullseries}
Z & = Z^{BP}\cdot z, &z & = \left( 1+\sum_{C\in\mathcal{C}}{r_C}\right),
& r_C & = \prod_{a\in C}{\mu_{a;\bar{a}_C}},
\end{align}
where $\mathcal{C}$ is the set of all the loops within the graph.  Each loop term $r_C$ is a
product of terms $\mu_{a,\bar{a}_C}$ associated with every node $a$
of the loop.  $\bar{a}_C$ denotes the set of neighbors of $a$ within the
loop $C$:
\begin{align}\label{eq:muterms}
\displaystyle
\mu_{a;\bar{a}_C} & = \displaystyle
\frac{\displaystyle \sum_{\boldsymbol{\sigma}_a}b_a\left(\boldsymbol{\sigma}_a\right)
{\displaystyle \prod_{b\in\bar{a}_C}{\left(\sigma_{ab}-m_{ab}\right)}}}
{\displaystyle \prod_{b\in\bar{a}_C}{ \sqrt{1-m_{ab}^2}}},
&
m_{ab} & = \sum_{\sigma_{ab}}{\sigma_{ab}b_{ab}\left(\sigma_{ab}\right)}.
\end{align}
In this work we consider planar graphs where all nodes are of degree not
larger than three, that is $|\bar{a}_C|\leq 3$.  We denote by \emph{triplet} a
node with degree three in the graph.  In Appendix A we show that a graphical
model can be converted to this representation at the cost of introducing
auxiliary nodes.
\begin{definition}\label{def:regloop}
A \textbf{2-regular loop} is a loop in which all nodes have degree \emph{exactly} two.
\end{definition}
\begin{definition}\label{def:regZ}
The \textbf{2-regular partition function} $Z_\emptyset$
is the truncated form of \eqref{eq:fullseries} which sums all $2$-regular loops only:
\footnote{Notice that this part of the series was called
\emph{single-connected partition function} in \citet{chertkov08}.  Here we prefer to
call it 2-regular partition function because loops with more than one connected
component are also included in this part of the series.}
\begin{align}\label{eq:disjseries}
Z_\emptyset & = Z^{BP}\cdot z_\emptyset, & z_\emptyset = 1+\sum_{C\in\mathcal{C} s.t. |\bar{a}_C|=2,\forall a\in C}{r_C}.
\end{align}
\end{definition}
\begin{figure}[!t]
\begin{center}
  \includegraphics[width=.8\columnwidth]{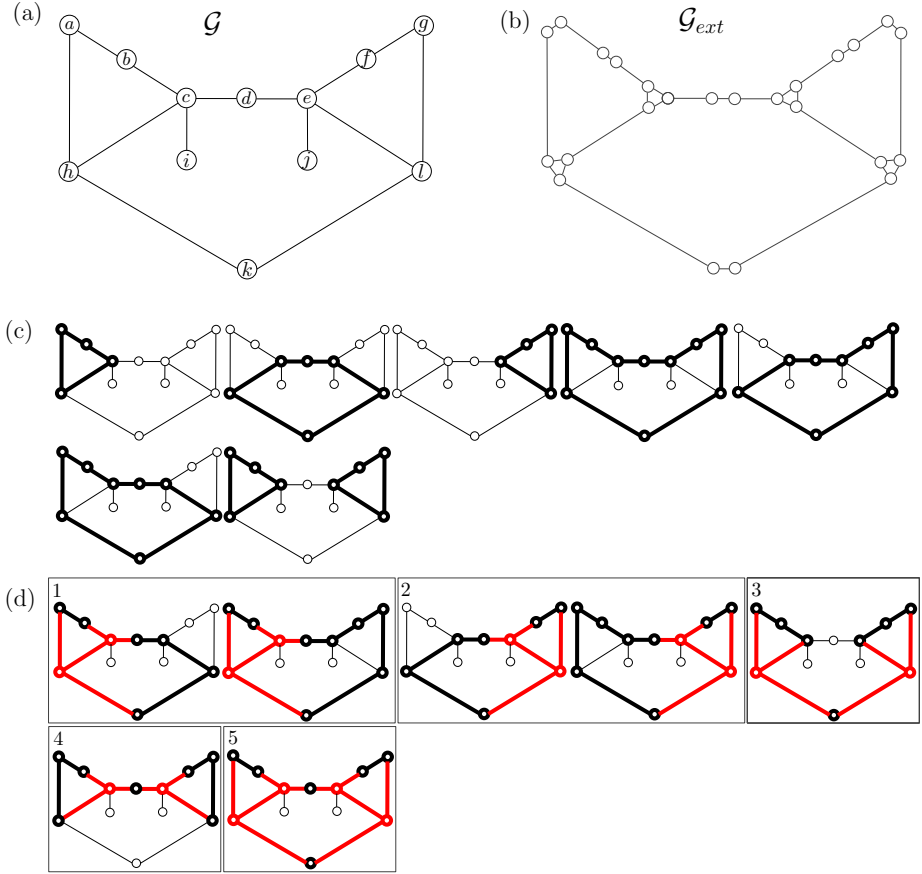}
\end{center}
\caption[Example]
{
Example. \textbf{(a)} A Forney graph. \textbf{(b)} Corresponding extended graph.
\textbf{(c)} Loops (in bold) included in the $2$-regular partition function.
\textbf{(d)} Loops (in bold and red) not included in the $2$-regular partition
function.  Marked in red, the triplets associated with each loop. Grouped in
gray squares, the loops considered in different subsets $\Psi$ of triplets:
(d.1) $\Psi=\{c,h\}$, (d.2) $\Psi = \{e,l\}$, (d.3) $\Psi = \{h,l\}$, (d.4)
$\Psi =\{c,e\}$ and (d.4) $\Psi=\{c,e,h,l\}$ (see
Section~\ref{sec:fullPfaffian}).  \label{fig:example}
}
\end{figure}
As an example, Figure \ref{fig:example}a shows a small Forney graph and Figure
\ref{fig:example}c shows seven loops found in the corresponding $2$-regular
partition function.
\subsection{Computing the $2$-regular Partition Function Using Perfect Matching}
\label{sec:disjoint}
In \cite{chertkov08} it has been shown that computation of $Z_\emptyset$
can be mapped to a dimer/matching problem, or equivalently, to the computation
of the sum of all weighted perfect matchings within another graph. A perfect matching is
a subset of edges such that each node neighbors exactly one edge from the
subset.  The weight of a matching is the product of weights of edges in the
matching.  The key idea of this mapping is to extend the original Forney graph
$\mathcal{G}$ into an new graph
$\mathcal{G}_{ext}:=(\mathcal{V}_{\mathcal{G}_{ext}},\mathcal{E}_{\mathcal{G}_{ext}})$
in such a way that each perfect matching in
$\mathcal{G}_{ext}$ corresponds to a $2$-regular loop in $\mathcal{G}$. (See Figures
\ref{fig:example}b and c for an illustration).  Under the condition of planarity, the sum of all weighted
perfect matchings can be calculated in a polynomial time following Kasteleyn's
arguments.  Here we reproduce these results with little variations and more
emphasis on the algorithmic aspects.

\begin{figure}[!t]
\begin{center}
  \includegraphics[width=.6\columnwidth]{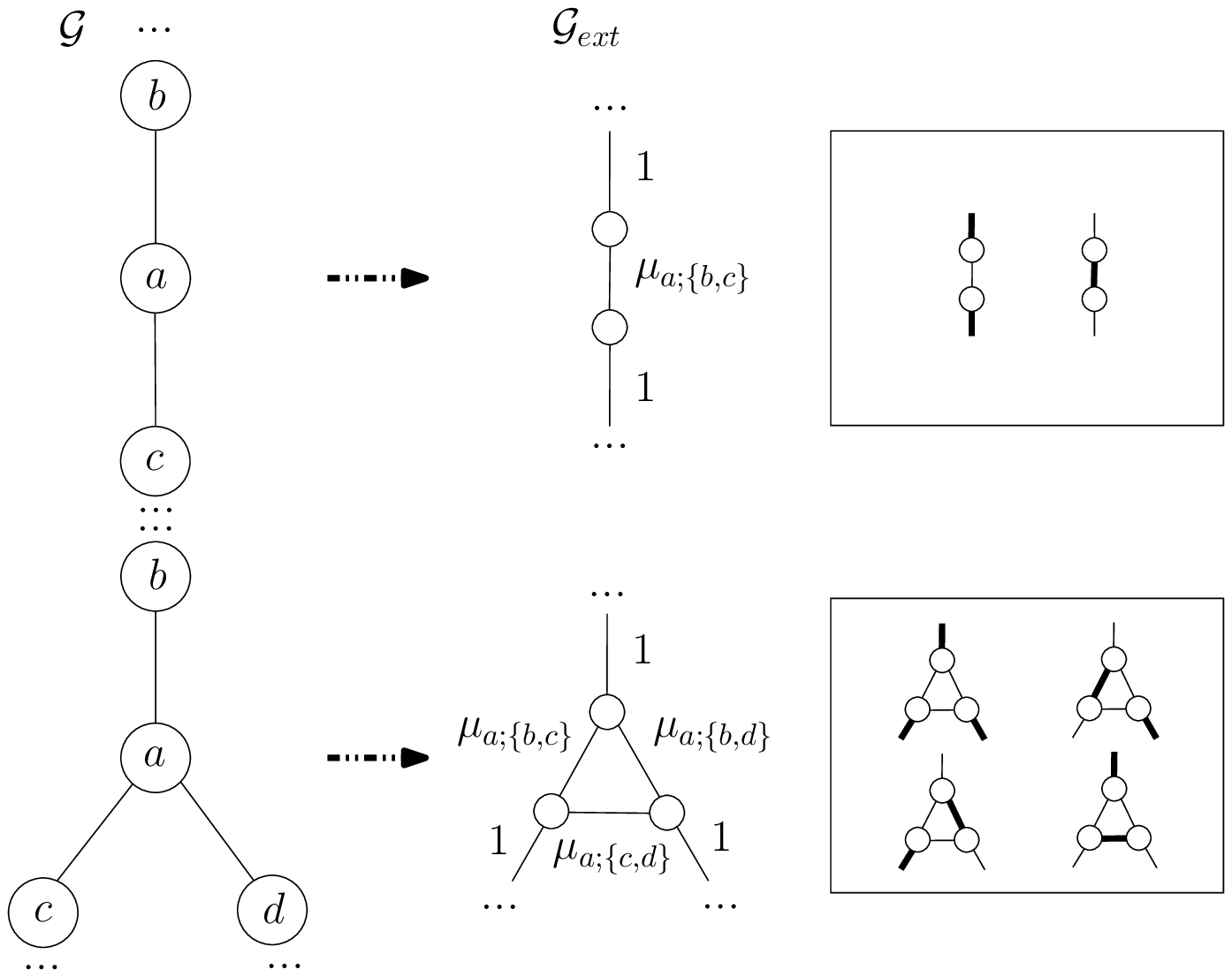}
\end{center}
\caption[Fisher's rules]
{
Fisher's rules.
\textbf{(Top)} A node $a$ of degree two in $\mathcal{G}$ is split in two nodes
in $\mathcal{G}_{ext}$.  \textbf{(Bottom)} A node $a$ of degree three in
$\mathcal{G}$ is split in three nodes in $\mathcal{G}_{ext}$.  The squares on
the right indicate all possible matchings in $\mathcal{G}_{ext}$ related with
node $a$. Note that the rules preserve planarity.
\label{fig:fisher}
}
\end{figure}

Given a Forney graph $\mathcal{G}$ and the BP approximation, we simplify
$\mathcal{G}$ and obtain the $2$-core by removing nodes of degree one
recursively.  After this step,  $\mathcal{G}$ is either the null graph (and then
BP is exact) or it is only composed of vertices of degree two or
three.

To construct the extended graph $\mathcal{G}_{ext}$ we split each node in
$\mathcal{G}$ according to the rules introduced by \citet{Fisher66} and
illustrated in Figure~\ref{fig:fisher}.  The procedure results in an extended graph of
$|\mathcal{V}_{\mathcal{G}_{ext}}|\leq 3|\mathcal{V}|$ nodes and
$|\mathcal{E}_{\mathcal{G}_{ext}}|\leq 3|\mathcal{E}|$ edges.  It is easy to
verify that each $2$-regular loop in $\mathcal{G}$ is associated with a perfect
matching in $\mathcal{G}_{ext}$ and, furthermore, this correspondence is
\emph{unique}.  Consider, for instance, the vertex of degree three in the
bottom of Figure~\ref{fig:fisher}.  Given a $2$-regular loop $C$, vertex $a$
can appear in four different configurations: either node $a$ does not appear in
$C$, or $C$ contains one of the following three paths: -$b$-$a$-$c$-,
-$b$-$a$-$d$- or -$c$-$a$-$d$-.  These four cases correspond to node terms in a
loop with values $1$, $\mu_{a;\{b,c\}}$, $\mu_{a;\{b,d\}}$ and
$\mu_{a;\{c,d\}}$ respectively, and coincide with the matchings in
$\mathcal{G}_{ext}$ shown within the box on the bottom-right.  An simpler
argument applies to the vertex of degree two from the top portion of
Figure~\ref{fig:fisher}.

Therefore, if we associate to each internal edge (new edge in
$\mathcal{G}_{ext}$ not in $\mathcal{G}$) of each split node $a$ the
corresponding term $\mu_{a;\bar{a}_C}$ of Equation \eqref{eq:muterms} and to
the external edges (existing edges already in $\mathcal{G}$) weight  $1$,
then the sum over all weighted perfect matchings defined on $\mathcal{G}_{ext}$
is precisely $z_{\emptyset}$.  The 2-regular partition function $Z_{\emptyset}$
is obtained using Equation \eqref{eq:disjseries}. Equivalently:
\begin{align}
z_\emptyset = \sum{\text{perfect matchings in $\mathcal{G}_{ext}$}}.\notag
\end{align}

\citet{kasteleyn63} provided a method to compute this sum in polynomial time
for planar graphs. We follow his approach.
First, we create a \emph{planar embedding} of $\mathcal{G}_{ext}$.
A planar embedding of a graph divides the plane into disjoint
regions that are bounded by sequences of edges in the graph. The regions are
called \emph{faces}.
Second, we orient the edges of the planar embedding in such a way that for
every face (except possibly the unbounded or external face) the number
of clockwise oriented edges is odd.
Algorithm \ref{al:Pfaffian-orientation} produces such an orientation
\citep{Karpinski98}. 
It receives an undirected graph $\mathcal{G}_{ext}$ and constructs a copy
$\mathcal{G}'_{ext}:=(\mathcal{V}_{\mathcal{G}'_{ext}},\mathcal{E}_{\mathcal{G}'_{ext}})$
with properly oriented edges $\mathcal{E}_{\mathcal{G}'_{ext}}$.  

%

It is convenient that $\mathcal{G}_{ext}$ is bi-connected, i.e. it has no
articulation points.  If needed, we add dummy edges with zero weight which do
not alter the partition function or the original model.

\floatname{algorithm}{Algorithm}
\begin{algorithm}
\caption{Pfaffian orientation}
\label{al:Pfaffian-orientation}
\begin{algorithmic}[1]
\REQUIRE undirected bi-connected extended graph $\mathcal{G}_{ext}$.\\
	\STATE Construct a planar embedding $\mathcal{\bar{G}}_{ext}$ of $\mathcal{G}_{ext}$.
	\STATE Construct a spanning tree $T$ of $\mathcal{\bar{G}}_{ext}$.
	\STATE Construct a graph $H$ having vertices corresponding to the faces of $\mathcal{\bar{G}}_{ext}$:\\
				Connect two vertices in $H$ if the respective face boundaries share an edge not in $T$.\\
				$H$ is a tree. Root $H$ to the external face.
	\STATE $\mathcal{G}'_{ext} := T$.
	\STATE Orient all edges in $\mathcal{G}'_{ext}$ arbitrarily.
  \FORALL{face (vertex in $H$) traversed in post-order}
		\STATE Add to $\mathcal{G}'_{ext}$ the unique edge not in $\mathcal{G}'_{ext}$.
		\STATE Orient it such that the number of clock-wise oriented edges is odd.
	\ENDFOR
\STATE \textbf{RETURN} directed bi-connected extended graph $\mathcal{G}'_{ext}$.
\end{algorithmic}
\end{algorithm}

Finally, denote $\mu_{ij}$ the weight of the edge between nodes $i$ and $j$ in
$\mathcal{G}'_{ext}$. We create the following skew-symmetric matrix $\hat{A} =
-\hat{A}^t$:
\begin{align*}
\hat{A}_{ij} & =
	\begin{cases}
		+\mu_{ij} & \text{if $(i,j)\in\mathcal{E}_{\mathcal{G}'_{ext}}$} \\
		-\mu_{ij} & \text{if $(j,i)\in\mathcal{E}_{\mathcal{G}'_{ext}}$} \\
		0         & \text{otherwise}
	\end{cases}.
\end{align*}

This matrix is known as the Tutte matrix of $\mathcal{G}'_{ext}$ and the
Pfaffian of $\hat{A}$ gives the desired sum up to the overall sign.  The
Pfaffian of $\hat{A} =\pm \sqrt{\text{Det}(\hat{A})}$.  However, $z_\emptyset$
can be either positive or negative, and computing the value of the Pfaffian
with the sign yet uncertain is not sufficient.  Furthermore, since each element
$\hat{A}_{ij}$ can be negative not only due to the Pfaffian orientation but
also if $\mu_{ij}$ is negative, the sign of the Pfaffian needs to be
\emph{corrected}.  This problem is fixed with the help of the original
Kasteleyn's binary matrix:
\begin{align*}
\hat{B}_{ij} & =
	\begin{cases}
		+1 & \text{if $(i,j)\in\mathcal{E}_{\mathcal{G}'_{ext}}$} \\
		-1 & \text{if $(j,i)\in\mathcal{E}_{\mathcal{G}'_{ext}}$} \\
		0         & \text{otherwise}
	\end{cases}.
\end{align*}

If the sign of $\text{Pf}(\hat{B})$ is negative then the sign of
$\text{Pf}(\hat{A})$ is changed.  Notice  that the absolute value of
$\text{Pf}(\hat{B})$ coincides with the number of perfect matchings or the
number of loops included in the sum if no additional edges have been added.
The sign of $\text{Pf}(\hat{B})$ represents the correction. Therefore, the
corrected value of $z_\emptyset$ is:
\begin{align*}
z_\emptyset & = \text{sign}\left(\text{Pf}\left(\hat{B}\right)\right)
\cdot \text{Pf}\left(\hat{A}\right).
\end{align*}

Calculation of $z_\emptyset$ can therefore be performed in time
$\mathcal{O}(N^3)$ where $N$ is the number of nodes of $\mathcal{G}_{ext}$
\citep{Galbiati94}.
For the special case of binary planar graphs with zero local fields the
$2$-regular partition function coincides with the exact partition function $Z =
Z_\emptyset = Z^{BP}\cdot z_\emptyset$ since the other terms in the loop series
vanish. 
\subsection{Computing the Full Loop Series Using Perfect Matching}
\label{sec:fullPfaffian}
\citet{chertkov08} established that $z_\emptyset$ is just the first term of a
finite sum involving Pfaffians. We briefly reproduce their results here and
provide an algorithm for computing the full loop series as a Pfaffian series.

Consider $\mathcal{T}$ defined as the set of all possible triplets (vertices with degree
three in the original graph $\mathcal{G}$).  For each possible subset $\Psi \in
\mathcal{T}$, including an \emph{even} number of triplets, there exists a
unique correspondence between  loops in $\mathcal{G}$ including the triplets in
$\Psi$ and perfect matchings in another extended graph
$\mathcal{G}_{{ext}_{\Psi}}$ constructed after removal of the triplets $\Psi$
in $\mathcal{G}$.  Using this representation the full loop series can be
represented as a Pfaffian series, where each term~$Z_{\Psi}$ is tractable and
is a product of the respective Pfaffian and the $\mu_{a;\bar{a}}$ terms
associated with each triplet of $\Psi$:
\footnote{
We omit the loop index in the triplet term $\mu_{a;\bar{a}}$ because nodes have
at most degree three and therefore the set $\bar{a}$ always coincide in
all loops which contain that triplet.
}
\begin{align}\label{eq:fullPfaffianseries}
z & = \sum_{\Psi}{Z_{\Psi}} & Z_{\Psi} & = z_{\Psi}\prod_{a\in\Psi}{\mu_{a;\bar{a}}} \\
& &z_{\Psi} & = \text{sign}\left(\text{Pf}\left(\hat{B}_{\Psi}\right)\right) \cdot \text{Pf}\left(\hat{A}_{\Psi}\right).\notag
\end{align}

The $2$-regular partition function thus corresponds to $\Psi = \emptyset$.  We
refer to the remaining terms of the series as higher order Pfaffian terms.
Notice that matrices $\hat{A}_{\Psi}$ and $\hat{B}_{\Psi}$ depend on the
removed triplets and therefore each $z_\Psi$ requires different matrices and
different edge orientations.  In addition, after removal of vertices in
$\mathcal{G}$ the resulting graph may be disconnected. As before, in these
cases we add dummy edges to $\mathcal{G}_{ext}$ with zero weight to make the
graph bi-connected again.  

Figure \ref{fig:example}d shows loops corresponding to the higher order
Pfaffian terms on our illustrative example. The first and second subsets of
triplets (d.1 and d.2) include summation over two loops whereas the
remaining Pfaffian terms include uniquely one loop.

Exhaustive enumeration of all the subsets of triplets leads to a
$2^{|\mathcal{T}|}$ time algorithm, which is prohibitive. However, many triplet
combinations may lead to forbidden configurations. Experimentally, we found
that a principled way to look for higher order Pfaffian terms with large
contribution is to search first for pairs of triplets, then groups of four, and
so on.  For large graphs, this also becomes intractable. Actually, the problem
is very similar to the problem of selecting loop terms $r_C$ with largest
contribution. The advantage of the Pfaffian representation, however, is that
$Z_\emptyset$ is always the Pfaffian term that accounts for the largest number
of loop terms and is the most contributing term in the series.  In this work we
do not derive any heuristic for searching Pfaffian terms with larger
contributions. Instead, in Section \ref{sec:exp-full} we study the full
Pfaffian series and subsequently we restrict ourselves on the accuracy of
$Z_\emptyset$.


Algorithm \ref{al:disjseries} describes the full procedure to compute all terms
using the representation of expression \eqref{eq:fullPfaffianseries}.  The main
loop of the algorithm can be interrupted at any time, thus leading to a sequence of
algorithms producing corrections incrementally.

\floatname{algorithm}{Algorithm}
\begin{algorithm}
\caption{Pfaffian series}
\label{al:disjseries}
\begin{algorithmic}[1]
\REQUIRE Forney graph $\mathcal{G}$\\
\STATE $z := 0$.
\FORALL{($\Psi \in \mathcal{T}$)}
	\STATE Build extended graph $\mathcal{G}_{{ext}_{\Psi}}$ applying rules of Figure \ref{fig:fisher}.
	\STATE Set Pfaffian orientation in $\mathcal{G}_{{ext}_{\Psi}}$ according to Algorithm \ref{al:Pfaffian-orientation}
	\STATE Build matrices $\hat{A}$ and $\hat{B}$.
	\STATE Compute Pfaffian with sign correction $z_{\Psi}$ according to Equation~\eqref{eq:fullseries}.
	\STATE $z := z + z_{\Psi}\prod_{a\in\Psi}{\mu_{a;\bar{a}}}$.
\ENDFOR
\STATE \textbf{RETURN} {$Z^{BP}\cdot z$}
\end{algorithmic}
\end{algorithm}

\section{Experiments}
\label{sec:experim}
In this Section we study numerically the proposed algorithm.  To facilitate the
evaluation and the comparison with other algorithms we focus on the binary
Ising model, a particular case of the model \eqref{eq:model} where factors only
depend on the disagreement between two variables and take the form
$f_a\left(\sigma_{ab},\sigma_{ac}\right) =
\exp\left(J_{a;\{ab,ac\}}\sigma_{ab}\sigma_{ac}\right)$.  We consider also
nonzero local potentials parametrized by $f_a\left(\sigma_{ab}\right) =
\exp\left(J_{a;\{ab\}}\sigma_{ab}\right)$ in all variables so that the model
becomes planar-intractable.

We create different inference problems by choosing different interactions
$\{J_{a;\{ab,ac\}}\}$ and local field parameters $\{J_{a;\{ab\}}\}$.  To
generate them we draw independent samples from a Normal distribution
$\{J_{a;\{ab,ac\}}\}	\sim \mathcal{N}(0, \beta/2)$ and $\{J_{a;\{ab\}}\} \sim
\mathcal{N}(0, \beta\Theta)$, where $\Theta$ and $\beta$ determine how
difficult the inference problem is.  Generally, for $\Theta = 0$ the planar
problem is tractable. For $\Theta > 0$, small values of $\beta$ result in
weakly coupled variables (easy problems) and large values of $\beta$ in
strongly coupled variables (hard problems).  Larger values of $\Theta$ result
in easier inference problems.

In the next Subsection we analyze the full Pfaffian series using a small
example and compare it with the original representation based on the loop
series.  Next, we compare our algorithm with the following ones:
\footnote{We use the libDAI library \citep{Moo08} for algorithms
\textbf{CVM-Loopk}, \textbf{TreeEP} and \textbf{TRW}.}
\begin{description}
\item[Truncated Loop-Series for BP] (TLSBP) \citep{gomez07}, which accounts for
a certain number of loops by performing depth-first-search on the factor graph
and then merging the found loops iteratively.  We adapted TSLBP as an any-time
algorithm (\textbf{anyTLSBP}) such that the length of the loop is used as the
only parameter instead of the two parameters $S$ and $M$ (see \citet{gomez07}
for details). This is equivalent to setting $M=0$ and discard $S$.  In this
way, anyTLSBP does not compute all possible loops of a certain length (in
particular, complex loops \footnote{A complex loop is defined as a loop which
can not be expressed as the union of two or more circuits or simple loops.} are
not included), but is more efficient than TLSBP.

\item[Cluster Variation Method] (\textbf{CVM-Loopk}) A double-loop
implementation of CVM \citep{hes03a}.  This algorithm is a special case of
generalized belief propagation \citep{yedidia05} with convergence guarantees.
We use as outer clusters all (maximal) factors together with loops of four
(k=4) or six (k=6) variables in the factor graph.

\item[Tree-Structured Expectation Propagation] (\textbf{TreeEP})
\citep{NIPS2003_AA25}.
This method performs exact inference on a base tree of the graphical model and
approximates the other interactions.  The method yields good results if the
graphical model is very sparse.
\end{description}
When possible, we also compare with the following two variational methods which
provide upper bounds on the partition function:
\begin{description}
\item[Tree Reweighting] (\textbf{TRW}) \citep{wainwright05} 
which decomposes the parametrization of a probabilistic graphical
model as a mixture of spanning trees of the model, and then uses the convexity
of the partition function to get an upper bound. 
\item[Planar graph decomposition] (\textbf{PDC}) \citep{Globerson2006}
which decomposes the parametrization of a probabilistic graphical
model as a mixture of tractable planar graphs (with zero
local field).
\end{description}
To evaluate the accuracy of the approximations we consider errors in $Z$ and,
when possible, computational cost as well.  As shown in \citet{gomez07}, errors
in $Z$, obtained from a truncated form of the loop series, are very similar to
errors in single variable marginal probabilities, which can be obtained by
conditioning over the variables under interest.  We only consider tractable
instances for which $Z$ can be computed via the junction tree algorithm
\citep{lauritzen88} using $8$GB of memory. When studying the scalability of the
approaches, we Given an approximation $Z'$ of $Z$, the error measure used in
this manuscript is:
\begin{align*}
\text{error} Z' & = \frac{|\log Z - \log Z'|}{\log Z}.
\end{align*}

As in \citet{gomez07}, we use four different message updates for
BP: fixed and random sequential updates, parallel (or synchronous) updates, and
residual belief propagation (RBP), a method proposed by~\citet{Elidanal} which
selects the next message to be updated which has maximum \emph{residual}, a
quantity defined as an upper bound on the distance of the current messages from
the fixed point.  We report non-convergence when none of the previous methods
converged.  We report convergence at iteration $t$ when the maximum absolute
value of the updates of all the messages from iteration $t-1$ to $t$ is smaller
than a threshold $\vartheta = 10^{-14}$.

\subsection{Full Pfaffian Series}
\label{sec:exp-full}
\begin{figure}[!th]
\begin{center}
  \includegraphics[width=.4\columnwidth]{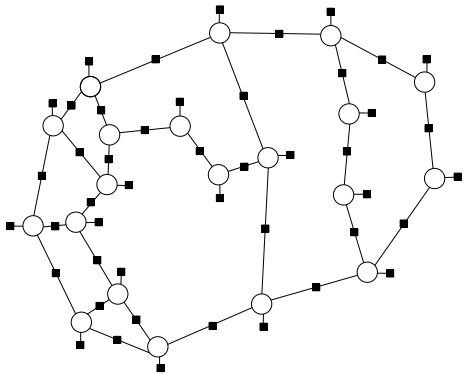}
\end{center}
\caption[example graph]
{
	Planar bipartite factor graph used to compare the full Pfaffian series with
	the loop series.  Circles and black squares denote	variables and factors
	respectively.
	\label{fig:example_graph}
}
\end{figure}

In the previous Section we have described two equivalent representations for the
exact partition function in terms of the loop series and the Pfaffian series.
Here we analyze numerically how these two representations differ using an
example, shown in Figure \ref{fig:example_graph} as a factor graph, for which
all terms of both series can be computed.  We analyze a single instance,
parametrized using $\Theta = 0.1$ and different pairwise interactions
$\beta\in\{0.1, 0.5, 1.5\}$. 
\begin{figure}[!t]
\begin{center}
  \includegraphics[angle=-90,width=.9\columnwidth]{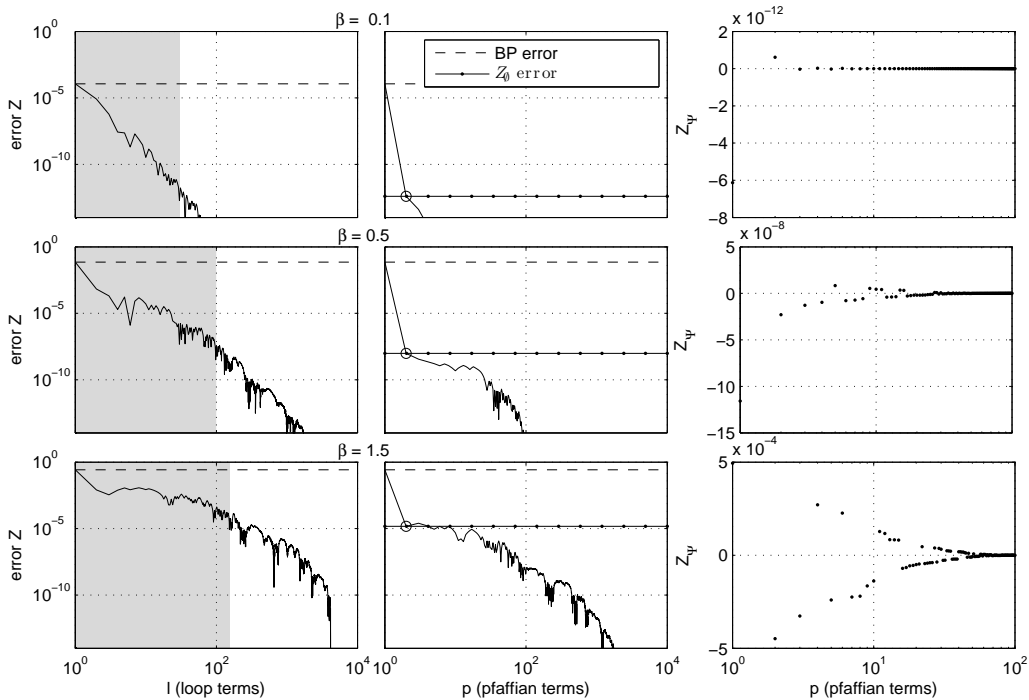}
\end{center}
\caption[example error]
{
Comparison between the full loop series and the full Pfaffian series.  Each row
corresponds to a different value of the interaction strength $\beta$.
\textbf{Left column} shows the error, considering loop terms $Z^{TLSBP}(l)$ in
log-log scale.  Shaded regions include all loop terms (not necessarily
$2$-regular loops) required to reach the same (or better) accuracy than the
accuracy of the $2$-regular partition function $Z_\emptyset$.  \textbf{Middle
column} shows the error considering Pfaffian terms $Z^{Pf}(p)$ also in log-log
scale. The first Pfaffian term corresponds to $Z_\emptyset$, marked by a
circle.  \textbf{Right column} shows the values of the first~$100$~Pfaffian
terms sorted in descending order in $|Z_{\Psi}|$ and excluding $z_\emptyset$.
\label{fig:example_error}
}
\end{figure}

We use TLSBP to retrieve all  loops, $8123$ for this example, and
Algorithm~\ref{al:disjseries} to compute all Pfaffian terms.  To compare the
two approximations we sort all contributions, either loops or Pfaffians, by
their absolute values in descending order, and then analyze how the errors are
corrected as more terms are included in the approximation.  We define partition
functions for the truncated series in the following way:
\begin{align}
Z^{TLSBP}(l) = & Z^{BP} \left( 1 + \sum_{i=1...l}{r_{C_i}} \right), &
Z^{Pf}(p) = & Z^{BP} \left( \sum_{i=1...p}{Z_{\Psi_i}} \right).\notag
\end{align}
Then $Z^{TLSBP}(l)$ accounts for the $l$ most contributing loops and
$Z^{Pf}(p)$ accounts for the $p$ most contributing Pfaffian terms.  In all
cases, the Pfaffian term with largest absolute value $Z_{\Psi_1}$ corresponds
to~$z_\emptyset$.

Figure~\ref{fig:example_error} shows the error $Z^{TLSBP}$ (first column) and
$Z^{Pf}$ (second column) for both representations.  For weak interactions
($\beta=0.1$) BP converges fast and provides an accurate approximation with an
error of order $10^{-4}$.  Summation of less than $50$ loop terms (top-left
panel) is enough to obtain machine precision accuracy. Notice that the error is
almost reduced totally with the $z_\emptyset$ correction (top-middle panel).
In this scenario, higher order terms of the Pfaffian series are negligible
(top-right panel).

As we increase $\beta$, the quality of the BP approximation decreases.  The
number of loop corrections required to correct the BP error then increases.  In
this example, for intermediate interactions ($\beta = 0.5$) the first Pfaffian
term $z_\emptyset$ improves considerably, more than five orders of magnitude,
on the BP error, whereas approximately $100$ loop terms are required to achieve
a similar correction (gray region of middle-left panel).

For strong interactions ($\beta = 1.5$) BP converges after many iterations and
gives a poor approximation. In this scenario also a larger proportion of loop
terms (bottom-left panel) is necessary to correct the BP result up to machine
precision.  Looking at the bottom-left panel we find  that approximately $200$
loop terms are required to achieve the same correction as the one obtained by
$z_\emptyset$. The $z_\emptyset$ is quite accurate (bottom-middle panel).

As the right panels show, higher order Pfaffian contributions change
progressively from a flat sequence of small terms to an alternating sequence of
positive and negative terms which grow in absolute value as $\beta$ increases.
These oscillations are also present in the loop series expansion.


In general, we conclude that the $z_\emptyset$ correction to the BP
approximation can give a significant improvement even in hard problems for which
BP converges after many iterations. Notice again that calculating
$z_\emptyset$, the most contributing term in the Pfaffian series, does not
require explicit search of loop or Pfaffian terms.


\subsection{Grids}
After analyzing the full Pfaffian series on a small random example we now
restrict our attention to the $Z_\emptyset$ approximation using Ising grids
(nearest neighbor connectivity).  First, we compare that approximation with
other inference methods for different types of interactions (attractive or
mixed) and then study the scalability of the method in the size of the
graphs.
\subsubsection{Attractive Interactions}
We first focus on binary models with interactions that tend to align the
neighboring variables to the same value, $J_{a;\{ab,ac\}} > 0$.  If local fields
are also positive $J_{a;\{ab\}} > 0, \forall{a\in\mathcal{V}}$,
\citet{sudderth2007} showed that, under some additional condition, the BP
approximation is a \emph{lower-bound} of the exact partition function and all
loops (and therefore Pfaffian terms too) have the same sign\footnote{The
condition is that all single variable beliefs at the BP fixed point must
satisfy $m_{ab} = b_{ab}(+1) - b_{ab}(-1) > 0, \forall{(a,b)\in\mathcal{E}}$}.
Although this is not formally proved for general models with attractive
interactions regardless of the sign of the local fields, numerical results suggest
that this property holds as well for this type of models.


We generate grids with positive interactions and local fields, that is
$|\{J_{a;bc}\}|	\sim \mathcal{N}(0, \beta/2)$ and $|\{J_{a;b}\}| \sim
\mathcal{N}(0, \beta\Theta)$, and study the performance for various values of
$\beta$, as well as for strong $\Theta = 1$ and weak $\Theta = 0.1$ local fields.

\begin{figure}[!t]
\begin{center}
  \includegraphics[angle=-90,width=\columnwidth]{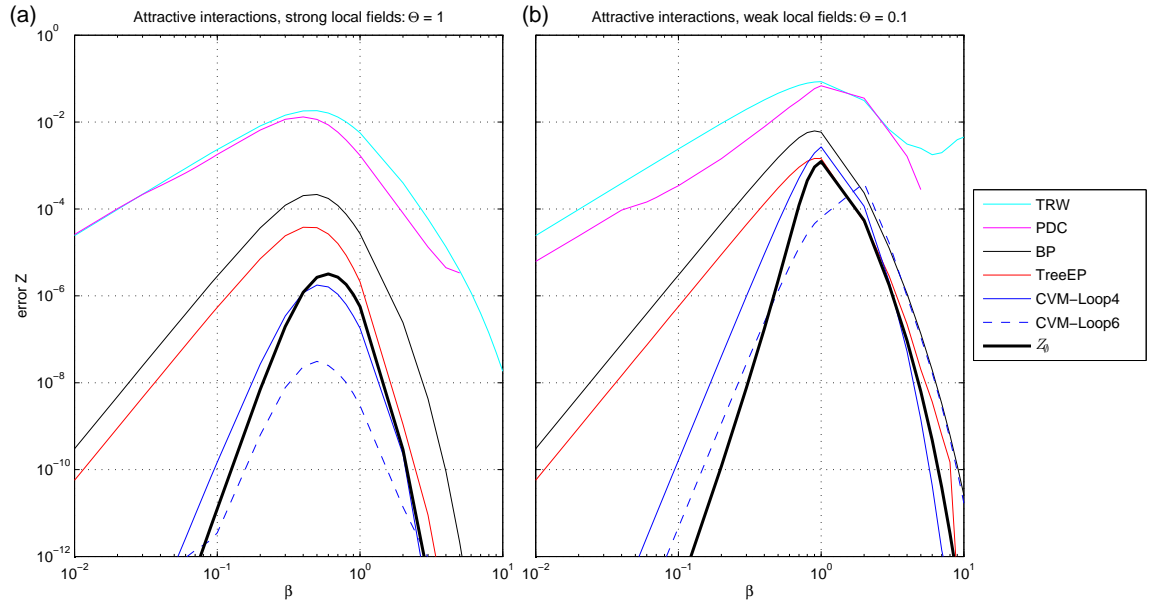}
\end{center}
\caption[7x7 grid attractive interactions]
{
7x7 grid attractive interactions and positive local fields.  Error averages
over $50$ random instances in function of the difficulty of the problem.
\textbf{(a)} Strong local fields.  \textbf{(b)} Weak local fields.
\label{fig:ising7x7_fe}
}
\end{figure}

Figure \ref{fig:ising7x7_fe} shows the average error over $50$ instances
reported by different methods. Using this setup, BP converged in all instances
using sequential updates of the messages.
The error curves of all methods show an initial growth and a subsequent
decrease, a fact explained by the phase transition occurring in this model for
$\Theta =0$ and $\beta \approx 1$ \citep{mooij2005}.  As the difference between
the two plots suggest, errors are larger as~$\Theta$ approaches zero. Notice,
that $Z_\emptyset=Z$ for the limit case of $\Theta=0$.

We observe that in all instances $Z_\emptyset$ \emph{always improves} over the BP
approximation.  Corrections are most significant for weak interactions
$\beta<1$ and strong local fields. For strong interactions $\beta>1$ and weak
local fields the improvement is less significant.

It appears that the $Z_\emptyset$ approximation performs better than TreeEP in all
cases except for very strong couplings, where they show very similar results.
Interestingly, $Z_\emptyset$ performs very similar to CVM-Loop4 which is known
to be a very accurate approximation for this type of model, see
\citet{yedidia00} for instance.  We observe that in order to obtain better
average results than $Z_\emptyset$ using CVM, we need to select larger
outer-clusters such as loops of length $6$, which increases dramatically the
computational cost.

The methods which provide upper bounds on $Z$ (PDC and TRW) report the largest
average error.  PDC performs slightly better than TRW, as was shown in
\citet{Globerson2006} for the case of mixed interactions.  We remark that the
worse performance of PDC for stronger couplings and weak local fields might be
attributed to implementation artifacts, since for $\beta>4$ we have numerical
precision errors.  In general, both upper bounds are significantly less tight
than the lower bounds provided by BP and $Z_\emptyset$.

\subsubsection{Mixed Interactions}
We now analyze a more general Ising grid model where interactions and local
fields can have mixed signs. In that case, $Z^{BP}$ and $Z_\emptyset$ are no
longer lower bounds on $Z$ and loop terms can be positive or negative.
\begin{figure}[!t]
\begin{center}
  \includegraphics[angle=-90,width=\columnwidth]{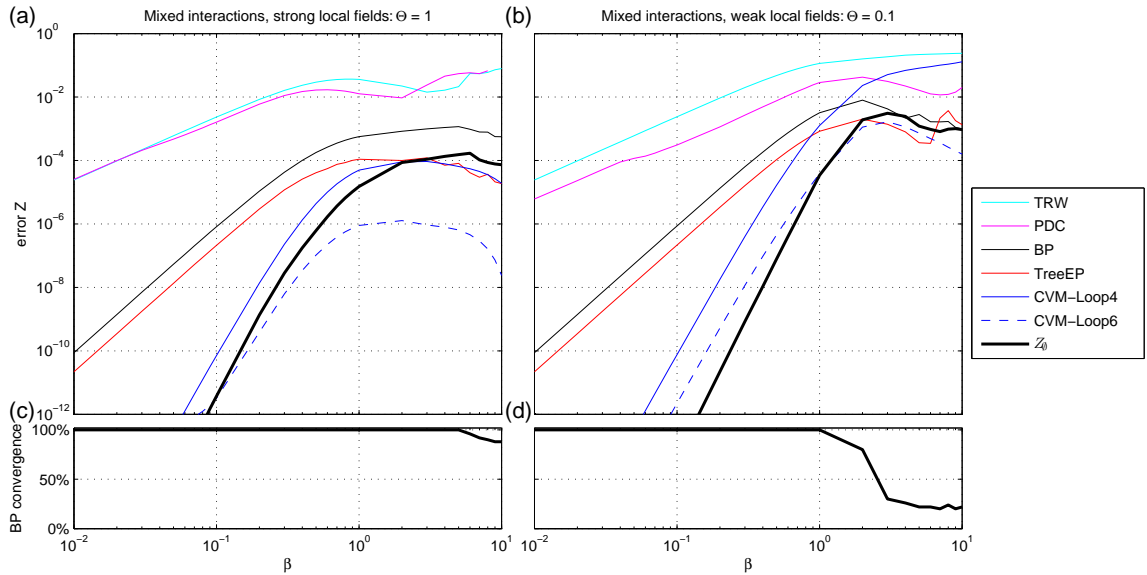}
	\end{center}
	\caption[7x7 grid mixed interactions]
{
7x7 grid mixed interactions. Error averaged over $50$ random
instances as a function of the problem difficulty for \textbf{(a)} strong local
fields and \textbf{(b)} weak local fields.
Bottom panels show percentage of cases when BP converges
for \textbf{(c)} strong local fields and \textbf{(d)} weak local fields.
\label{fig:ising7x7_sg}
}
\end{figure}
Figure \ref{fig:ising7x7_sg} shows results using this setup. Top panels show
average errors and bottom panels show percent of instances in which BP
converged using at least one of the methods described above.

For strong local fields (subplots a,c), we observe that $Z_\emptyset$
improvements over BP results become less significant as $\beta$ increases.
It is important to note that $Z_\emptyset$ always improves on the BP result,
even when the couplings are very strong  ($\beta = 10$) and BP fails to
converge for a small percentage of instances.  $Z_\emptyset$ performs slightly
better than CVM-Loop4 and substantially better than TreeEP for small and
intermediate $\beta$ .  All three methods show similar results for strong
couplings $\beta > 2$.  As in the case of attractive interactions, the best
results are attained using CVM-loop6.

For the case of weak local fields (subplots b,d), BP fails to converge near
the transition to the spin-glass phase.  For $\beta=10$, BP converges only in
less than $25\%$ of the instances.  In the most difficult domain, $\beta
>22$, all methods under consideration give similar results (all comparable to
BP). Moreover, it may happen that $Z_\emptyset$ degrades the $Z^{BP}$
approximation because loops of alternating signs have major influence in the
series. This result was also reported in~\citet{gomez07} when loop terms,
instead of Pfaffian terms, where considered.

\subsubsection{Scaling with Graph Size}
We now study how the accuracy of the $Z_\emptyset$ approximation changes as we
increase the size of the grid. We generate random grids with mixed couplings
for $\sqrt{N} = \{4, ..., 18\}$ and focus on a regime of very weak local fields
$\Theta = 0.01$ and strong couplings $\beta = 1$, a difficult configuration
according to the previous results.  We compare $Z_\emptyset$ also with
anyTLSBP, a variant of our previous algorithm for truncating the loop series.
We run anyTLSBP by selecting loops shorter than a given length, and the length
is increased progressively.  To provide a fair comparison between both methods,
we run anyTLSBP for the same amount of cpu time as the one required to obtain
$Z_\emptyset$.

\begin{figure}[!t]
\begin{center}
  \includegraphics[angle=-90,width=\columnwidth]{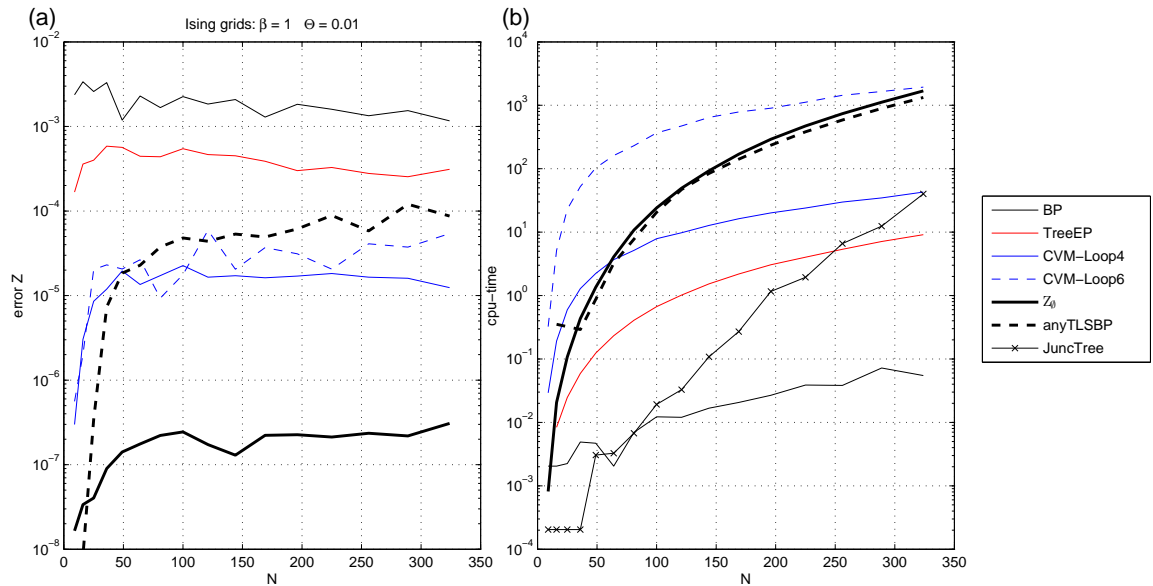}
	\end{center}
	\caption[Scaling median]
{
	Results on regular grids: scaling with grid size for strong interactions
	$\beta = 1$ and very weak local fields $\Theta = 0.01$.  BP converged in all
	cases.  \textbf{(a)} Error medians over $50$ instances.  \textbf{(b)} Cpu
	time (log-scale).
		\label{fig:scaling_median}
}
\end{figure}
Figure~\ref{fig:scaling_median}a shows the errors of different methods.  Since
variability in the errors is larger than before, we take the median for
comparison.  In order of increasing accuracy we get BP, TreeEP, anyTLSBP,
CVM-Loop6, CVM-Loop4 and $Z_\emptyset$.  We note that larger clusters in CVM
does not necessarily result in better performance.

Overall, we can see that results are roughly independent of the network size
$N$ in almost all methods that we compare. The error of anyTLSBP starts being
the smallest but soon increases because the proportion of loops captured
decreases very fast. For $N~>~64$, anyTLSBP performs worse than CVM.  The
$Z_\emptyset$ correction, on the other hand, stays flat and we can conclude
that it scales reasonably well.  Interestingly, although  $Z_\emptyset$ and
TLSBP use different ways to truncate the loop series, both methods show similar
scaling behavior for large graphs. 

%

Figure \ref{fig:scaling_median}b shows the cpu time for all the tested
approaches averaged over all cases.  Concerning the approximate inference
methods, in order of increasing cost, we have BP, TreeEP, CVM-Loop4,
$Z_\emptyset$ with anyTLSBP, and CVM-Loop6.  Although the cpu time required to
compute~$Z_\emptyset$~scales with $O(N_{G_{ext}}^3)$, its curve shows the
steepest growth. We discuss how to correct this caveat in Section
\ref{sec:discussion}. The cpu time of the junction tree method quickly
increases with the tree-width of the graphs.  For large enough~$N$, exact
solution via the junction tree method is no longer feasible because of its
memory requirements.  In contrast, for all approximate inference methods,
memory demands do not represent a limitation.

\subsection{Radial grid graphs}
\begin{figure}[!t]
\begin{center}
\includegraphics[angle=-90,width=.45\columnwidth]{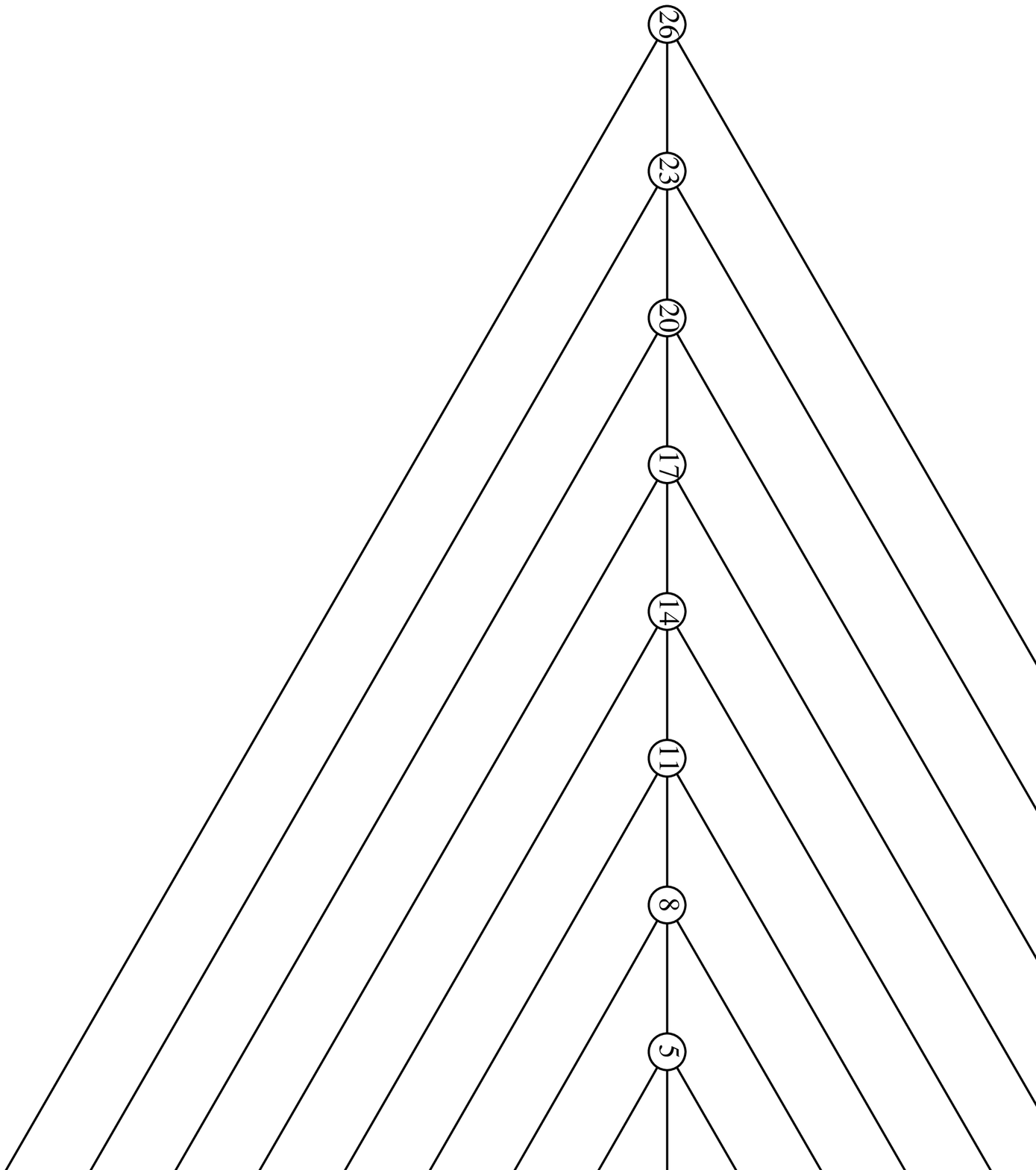}
\includegraphics[angle=-90,width=.45\columnwidth]{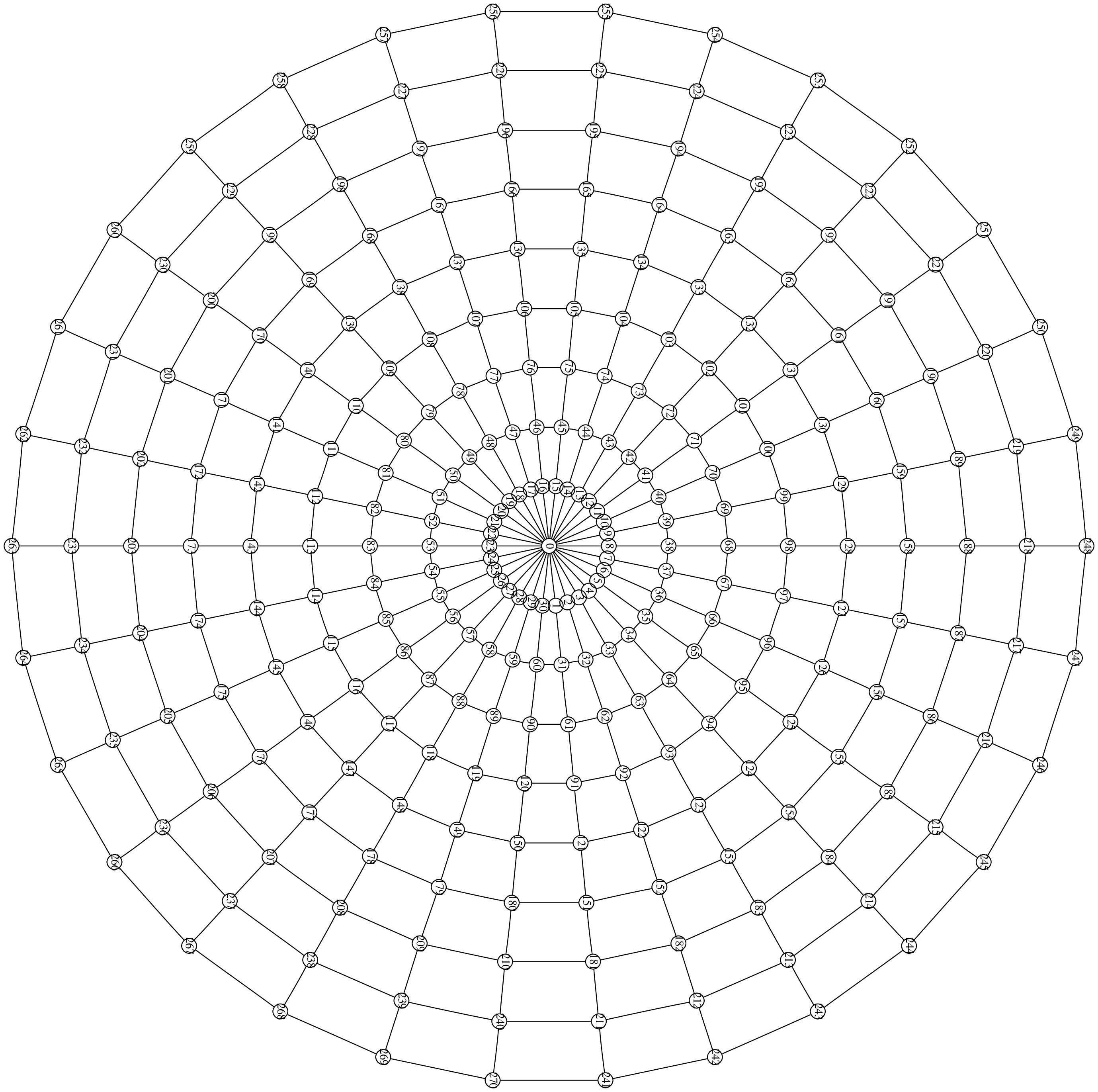}
\end{center}
\caption[Spider-web like graphs]
{
Two examples of planar graphs used for comparison between methods. We fix the
number of concentric polygons to $9$ and change the degree $d$ of the central
node within the range $[3,...,25]$.  \textbf{(left)} Graph for $d=3$.
\textbf{(right)} Graph for $d=25$.  Here nodes represent variables and edges
pairwise interactions.  We also add external fields which depend on
the state of each nodes (not drawn).
\label{fig:spide-web-graphs}
}
\end{figure}
In the previous subsection we analyzed the quality of the $Z_\emptyset$
correction for graphs with a regular grid structure.  Here, we carry over the
analysis of the $Z_\emptyset$ correction using planar graphs which consist of
concentric polygons with a variable number of sides. Figure
\ref{fig:spide-web-graphs} illustrates these spider-web like graphs.  We
generate them as factor graphs with pairwise interactions which we subsequently
convert to an equivalent Forney graph. (See Appendix A for details).  Again,
local field potentials are parametrized using $\Theta=0.01$ and interactions
using $\beta=1$. We analyze the error in $Z$ as a function of the degree $d$ of
the central node.

\begin{figure}[!t]
\begin{center}
\includegraphics[angle=-90,width=\columnwidth]{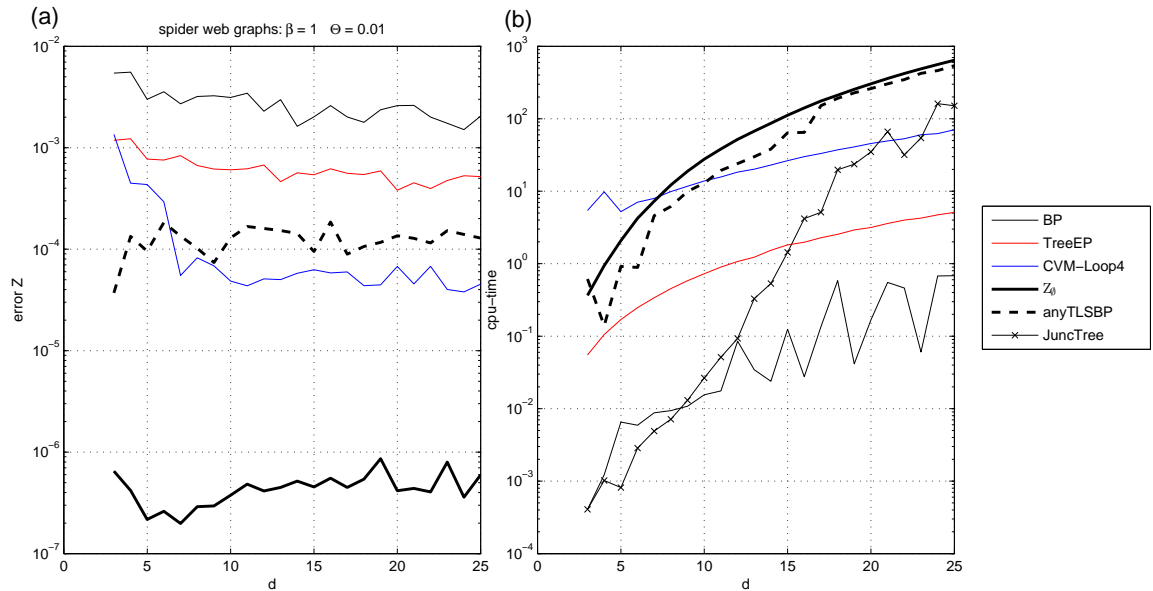}
\end{center}
\caption[Spider web graphs results.]
{
Results on spider-web like graphs: scaling with the degree of the central node
for $\beta = 1$ and $\Theta = 0.01$.  BP converged in all cases.  \textbf{(a)}
Error medians over $50$ instances.  \textbf{(b)} Cpu time (log-scale).
\label{fig:regular_weak_median}
}
\end{figure}

Figure \ref{fig:regular_weak_median}a shows the median of errors in $Z$ of $50$
random instances.  First, we see that the variability of all methods, in
particular anyTLSBP, is larger than in the regular grid scenario.  Also, the
improvement of CVM-Loop4 over BP is slightly less significant, possibly caused
by the existence of the central node with a large degree.
CVM-Loop6 does not converge for instances with $d>4$ before $10^4$ seconds and
is not included in the analysis.  We can say that all approaches present
results independent of the degree $d$.

The $Z_\emptyset$ approximation is the best method compared to the other
tested approaches. The improvements of $Z_\emptyset$ on CVM-Loop4 (the
second best method) can be of more than two orders of magnitude and more than
three orders of magnitude compared to BP.

Computational costs are shown in \ref{fig:regular_weak_median}b. The best
performance of $Z_\emptyset$ comes at the cost of being the most expensive
approximate inference approach for which we obtain results.  Again, for larger
graphs, exact solution via the junction tree is not feasible due to the large
tree-width.

\section{Discussion}\label{sec:discussion}
We have presented an approximate algorithm based on the exact loop calculus
framework for inference on planar graphical models defined in terms of binary
variables.  The proposed approach improves the estimate for the partition
function provided by BP without an explicit search of loops.

The algorithm is illustrated on the example of ordered and disordered Ising
model on a planar graph.  Performance of the method is analyzed in terms of its
dependence on the system size. The complexity of the partition function
computation is exponential in the general case, unless the local fields are
zero,  when it becomes polynomial.  We tested our algorithm on regular grids
and planar graphs with different structures. Our experiments on regular grids
show that significant improvements over BP are always obtained if single
variable potentials (local magnetic fields) are sufficiently large.  The
quality of this correction degrades with decrease in the amplitude of external
field, to become exact at zero external fields.  This suggests that the
difficulty of the inference task changes abruptly from very easy, with no local
fields, to very hard, with small local fields, and then decays again as
external fields become larger.

%


The $Z_\emptyset$ correction turns out to be competitive with
other state of the art methods for approximate inference of the partition
function.  First of all, we showed that $Z_\emptyset$ is much more
accurate than upper bounds based methods such as TRW or PDC.
This illustrates that such methods come at the cost of less accurate
approximations.  We have also shown that for the case of grids with attractive
interactions, the lower bound provided by $Z_\emptyset$ is the most accurate.

Secondly, we found that $Z_\emptyset$ performs much better than treeEP for weak
and intermediate couplings and shows competitive results for strong
interactions.  Concerning CVM, we showed that using larger outer clusters does
not necessarily lead to better approximations.  In general, the $Z_\emptyset$
correction presented better results than CVM for our choice of regions.



Finally, we have presented a comparison of $Z_\emptyset$ with TLSBP, which is
another algorithm for the BP-based loop series using the loop length as
truncation parameter.
On the one hand, the calculation of $Z_\emptyset$ involves a \emph{re-summation}
of many loop terms which in the case of TLSBP are summed individually.  This
consideration favors the $Z_\emptyset$ approach.  On the other hand,
$Z_\emptyset$ is restricted to the class of 2-regular loops whereas TLSBP also
accounts for terms corresponding to more complex loop structures in which nodes
can have degree larger than two.  Overall, for planar graphs, we have shown
evidence that the $Z_\emptyset$ approach is better than TLSBP when the size of
the graphs is not very small.  We emphasize, however, that TLSBP can be applied
to non-planar binary graphical models too.


Currently, the shortcoming of the presented approach is in its relatively
costly implementation.  However, since the bottleneck of the algorithm is the
Pfaffian calculation and not the algorithm itself (used to obtain the extended
graphs and the associated matrices), it is easy to devise more efficient
methods than the one used here.  Thus,  one may substitute brute-force
evaluation of the Pfaffians by a smarter one available for planar graphs. This
reduces the cost from $\mathcal{O}(N^3)$ to $\mathcal{O}(N^{3/2})$
\citep{galluccio, loh}.  Besides, the Pfaffian of $\hat{B}$ is binary, see Eq.
\eqref{eq:fullPfaffianseries}, making it possible to improve using a
bit-matrix representation \citep{Schraudolph}.  Alternatively one could think
of a strategy which does not require the Pfaffian of $\hat{B}$.  All these
technical issues are the focus of our continuing investigation.

In this manuscript we have focused on inference problems defined on planar
graphs with symmetric pairwise interactions and, to make the problems
difficult, we have introduced local field potentials.  Notice however, that the
algorithm can also be used to solve models with more complex interactions, i.e.
more than pairwise as in the case of the Ising model \citep[see][for a
discussion of possible generalizations]{chertkov08}. This makes our approach
more powerful than other approaches, namely, \citep{Globerson2006,
Schraudolph}, designed specifically for the pairwise interaction case.

Although planarity is a severe restriction, we emphasize that planar graphs
appear in many contexts such as computer vision and image processing, magnetic
and optical recording, or network routing and logistics.  It would also be
interesting (and possible) to consider extensions of the algorithm developed in
the manuscript for approximate inference of some class of non-planar graphs.
Thus, following the approach of \citet{Globerson2006}, one can think of other
types of spanning subgraphs more general than "easy" planar graphs for which
exact computation can be performed using perfect matching. The correction
$Z_\emptyset$ can be an accurate approximation for this spanning subgraphs and
the resulting approximation method would also provide bounds on the exact
result.



%
%

\acks{
We acknowledge J.~M.~Mooij for providing the libDAI framework and A.~Windsor
	for the planar graph functions of the boost graph library.  We also thank
		V.~Y.~Chernyak, J.~K.~Johnson and N.~Schraudolph for interesting
		discussions and A.~Globerson for providing the Matlab sources of PDC.
			This research is part of the Interactive Collaborative In-
			formation Systems (ICIS) project, supported by the Dutch
			Ministry of Economic Affairs, grant BSIK03024.
The work at LANL was carried out under
the auspices of the National Nuclear Security Administration
of the U.S. Department of Energy at Los Alamos National
Laboratory under Contract No. DE-AC52-06NA25396.
}









\appendix
\section*{Appendix A: Converting a factor graph to a Forney Graph.}
\label{app:convert}
A probabilistic model is usually represented as a Bayesian Network or a Markov
Random Field. Since bipartite factor graphs subsume both models, we show here
how to convert a factor graph model defined in terms of binary variables to a more
general Forney graph representation, for which the presented algorithm can be
directly applied to.

On a bipartite factor graph $\mathcal{G_{F}}=(\mathcal{V_{F}},\mathcal{E_{F}})$ the set
$\mathcal{V_{F}}$ is composed of a set of variable nodes $\mathcal{I}$
and a set of factor nodes $\mathcal{J}$.
Each variable node $i\in\mathcal{I}, i:=\{1,2,\dots\}$ represents a variable which takes values
$\sigma_i = \{\pm 1\}$.  We label factor nodes using capital letters so that
$a=\{A,B,\dots\}, a\in\mathcal{J}$ denotes a factor node which has an
associated function $f_a(\boldsymbol{\sigma}_a)$ defined on a subset of
variables $\bar{a}\in\mathcal{I}$.
An (undirected) edge exists between two nodes $(a,i)\in\mathcal{E_{F}}$ if $i\in\bar{a}$.

Given $\mathcal{G_{F}}$, a direct way to obtain an equivalent Forney graph
$\mathcal{G}$ is: first, to create a node $\delta_i\in\mathcal{V}$ for each
variable node $i\in\mathcal{V_{F}}$, and  second, to associate a new binary
variable $\delta_ia$ with values $\sigma_{\delta_ia} = \{\pm1\}$ to edges
$(\delta_i,a)\in\mathcal{E}$.  Nodes $\delta_i\in\mathcal{V}$ are
\emph{equivalent factor nodes} denoting the characteristic function:
$\delta_i(\boldsymbol{\sigma}_a)=1$ if $\sigma_{\delta_ia}=\sigma_{\delta_ib}$,
	$\forall a,b\in\bar{\delta_i}$ and zero otherwise.  Finally, factor nodes
	$c\in\mathcal{V_{F}}$ correspond to the same factor nodes $c$ in
	$\mathcal{V}$ but defined in terms of the new variables $\delta_ic$, $\forall
	i\in\bar{c}$.

\begin{figure}[!t]
\begin{center}
  \includegraphics[width=.8\columnwidth]{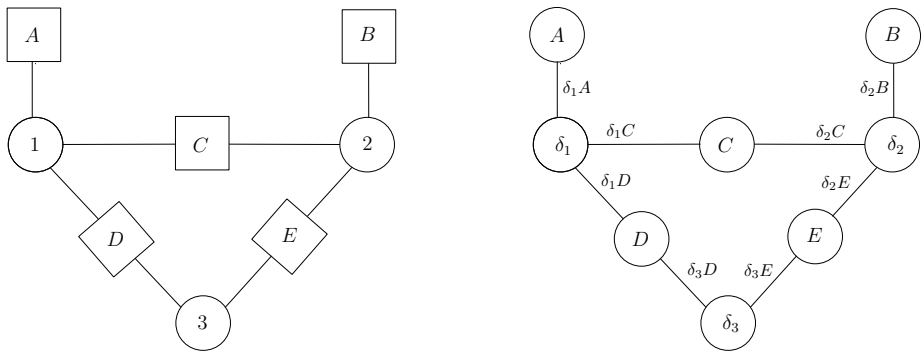}
\end{center}
\caption[Example]
{
\textbf{(a)} An factor graph $\mathcal{G_{F}}$ and \textbf{(b)} an equivalent Forney graph $\mathcal{G}$.
\label{fig:example-forney}
}
\end{figure}
Figure \ref{fig:example-forney} shows an example of this transformation.
Notice that, although we impose an direction in the edge labels, they remain
undirected: ($\delta_i,a$) = ($a,\delta_i$), $\forall \delta_i,a \in
\mathcal{V}$.  For variables $i\in\mathcal{V_{F}}$ which only appear in two
factors, such as variable $3$, the corresponding $\delta_3$ node is redundant
and can be removed.  The joint distribution of $\mathcal{G_F}$ is related to
the joint distribution of $\mathcal{G}$ by:
\begin{align}
&\frac{1}{Z}f_A(\sigma_1)f_B(\sigma_2)f_C(\sigma_1,\sigma_2)f_D(\sigma_1,\sigma_3)f_E(\sigma_2,\sigma_3) \\
\equiv& \frac{1}{Z}f_A(\sigma_{\delta_1A})f_B(\sigma_{\delta_2B})f_C(\sigma_{\delta_1C}, \sigma_{\delta_2C})
	f_D(\sigma_{\delta_1D}, \sigma_{\delta_3D})f_E(\sigma_{\delta_2E}, \sigma_{\delta_3E})\notag\\
&	f_{\delta_1}(\sigma_{\delta_1A}, \sigma_{\delta_1C}, \sigma_{\delta_1D}) f_{\delta_2}(\sigma_{\delta_2B}, \sigma_{\delta_2C},  \sigma_{\delta_2E})
	f_{\delta_3}(\sigma_{\delta_3D}, \sigma_{\delta_3E}).\notag
\end{align}
Once $\mathcal{G}$ has been generated following the previous procedure it may
be the case that the nodes $\delta_i\in\mathcal{V}$ have degree three or
larger.  This happens if a variable $i$ appears in more than $3$ factor nodes
on $\mathcal{G_F}$.  It is easy to convert $\mathcal{G}$ to a graph were all
$\delta_i$ nodes have maximum degree three by introducing new auxiliary
variables $\delta_{i_1}, \delta_{i_2}, ...$ and new equivalent nodes.  For
instance, if variable $i\in\mathcal{V_F}$ appears in $4$ factors $A,B,C,D$:
\begin{align}
f_{\delta_i}(\sigma_{\delta_iA},\sigma_{\delta_iB},\sigma_{\delta_iC},\sigma_{\delta_iD}) & \equiv
f_{\delta_{i_1}}(\sigma_{\delta_iA},\sigma_{\delta_iB},\sigma_{\delta_{i_1}})
f_{\delta_{i_2}}(\sigma_{\delta_{i_1}},\sigma_{\delta_iC},\sigma_{\delta_iD}).\notag
\end{align}

Notice that although the models are equivalent, the number of loops in
$\mathcal{G}$ may be larger than in $\mathcal{G_F}$.
In the case that a factor in $\mathcal{G_F}$ involves more than three
variables, as sketched in \citet{chertkov08}, one could split the node of
degree $N$ into auxiliary nodes of degree $N-1$ and compute $Z_\emptyset$ on the
transformed model. Alternatively, one can reduce the number of variables
that enter a factor by clamping.

\vskip 0.2in
\bibliographystyle{apalike}
\bibliography{gomez08a}

\end{document}